# Comprehensive Survey of Evolutionary Morphological Soft Robotic Systems

Reem J. Alattas*, Sarosh Patel, and Tarek M. Sobh

*Interdisciplinary Robotics, Intelligent Sensing and Control (RISC) Lab, School of Engineering, University of Bridgeport, 221 University Avenue, Bridgeport, CT 06604, USA*

* Correspondence should be addressed to Reem J. Alattas; ralataas@my.bridgeport.edu

# Comprehensive Survey of Evolutionary Morphological Soft Robotic Systems

Evolutionary robotics aims to automatically design autonomous adaptive morphological robots that can evolve to accomplish a specific task while adapting to environmental changes. Soft robotics have demonstrated the feasibility of evolutionary robotics for the synthesis of robots' control and morphology. The motivation of developing evolutionary soft computing techniques to that can generate task oriented structures for morphological robots makes the domain of soft-robotics worthy of serious investigation and research, and hence this article summarizes an important volume of research for a computational and software architecture perspective. This paper reviews the literature and discusses various aspects of evolutionary robotics including the application on morphological soft robots to allow self-assembly, self-reconfiguration, self-repair, and self-reproduction. Then, major milestones are outlined along with important morphological soft robotic prototypes due to their importance in the field. Finally, the current state of the art in the field is assessed.

## 1 Introduction

Producing autonomous adaptive robots is considered as a huge challenge. In biology, autonomous and adaptive creatures are produced using evolution. However, mainstream robots use machine learning to produce adaptive behaviour to simulate biological aspects, while neglecting the autonomous side of it. Therefore, evolutionary algorithms are used to optimize robots autonomy and adaptation producing what is known as evolutionary robots [1].

Evolutionary robotics approach evolves populations of simulated robots by synthesizing the robots' morphology and control using evolutionary computation methods, and then selects the fittest to be manufactured. The evolutionary approach continuously designs and builds different robots with improved capabilities rather than using the hand design approach which can be extremely difficult when designing autonomous adaptive robots. Thus far everything has a cost, and the cost in this case is the lack of guarantees that an optimal solution will be found, but the benefits of this method outweigh the cost. These benefits include the power of evolutionary algorithms to improve the parameters and the structure of the robots' control and morphology [2-3].

This paper starts by reviewing the literature of evolutionary robotic systems not in a chronological order, but in an order where each study is depending on the results of the previous studies to make more sense to the reader. Then, morphological soft robots are discussed as a method to implement evolutionary robots in the physical world, as advanced technologies and rapid prototyping techniques made these kinds of robots feasible. Moreover, evolutionary robotics can empower soft modular robots by allowing them to self-assemble, self-reconfigure, self-repair, and self-reproduce. Thereafter, we evaluate numerous soft modular robots applications and we analyse their capabilities of performing various evolvability challenges; i.e. self-assembly, self-reconfiguration, self-repair, and self-reproduction. Finally, we finish by outlining the current state of the art. Both of the modular robots and current state of the art literature is ordered chronologically according to the publication date.

## 2    Evolutionary Robotics

In nature, evolution produces heritable changes in organisms' phenotypes over multiple generations for better adaptation to the environment. In robotics, evolution was introduced as a nature inspired approach to avoid the bias and limitations introduced by human designers and to produce better adapted robots to the environmental changes [4]. Simply, evolutionary robotics can be considered as a method of creating autonomous robots automatically without human intervention [5].

Evolutionary robotics is inspired by the Darwinian theory of evolution which states that all organisms develop through mutation, crossover, and selection that increase the new generation's ability to compete, survive, and reproduce [6]. Based on the principle of selective reproduction of the fittest, robots are viewed as autonomous artificial organisms that can develop their own skills by interacting with the environment and without human intervention. The fittest robots survive and reproduce until a robot that satisfies the performance criteria is produced [7].

The literature below is not ordered chronologically, but in an order that contributes better to the milestones sequence, where each study depends on the results of the previous ones in this article.

Nolfi and Floreano presented a set of experiments in their book, ranging from simple to very complex, in order to address different adaptation mechanisms. The first set of experiments involves navigational tasks; such as obstacle avoidance. The authors point out that in some cases the evolved solution outperformed the hand-designed solution by capitalizing on interactions between machine and environment that could not be captured by a model based approach. On the other hand, more complex tasks expose limits of reactive architectures. However, very complex tasks such as garbage collection and battery recharging show that emergent modular structures allowed the decomposition of the global behaviour into basic behaviours to emerge spontaneously. Furthermore, the achieved decomposition did not correspond to a distal decomposition an external designer would naturally expect, and outperformed other manually designed decompositions [7].

Lipson stated that each robot comprises two major parts: controller (brain) and morphology (body). Controllers can be represented in many ways including neural networks that map sensory input to actuator outputs. Morphology can be described as tree-based representation, L-system consisting of set of rules that can produce construction sequences, or regulatory networks. To allow for open-ended synthesis, both controller and morphology should co-evolve along with the fitness functions and evaluation methods [8].

Floreano et al. described evolving a small wheeled robot's controller (neural network) using a simple genetic algorithm to navigate a looping maze. The experiment showed that the fitness function evolved and the cruising speed of the robot evolved as well, which demonstrates that evolution can lead to better adaptation [5].

Bongard explored the same concept on a legged robot in a physically realistic simulator. The goal of the experiment was to evolve the controller (neural network) to make the robot locomote towards the high chemical concentration area. The resulting robot moved and changed direction towards the high concentration areas, which shows that two independent functions evolved successfully; locomotion and gradient tracking [2].

Zykov et al. applied the same theory on a physical robot to evolve the dynamic gates in hardware. The nine-legged robot's open-loop controller was evolved using a genetic algorithm to allow evolving speed and locomotion pattern under the rhythmicity constraint [9].

Paul and Bongard introduced coupled evolution of robotic morphology and control on a biped robot in simulation. The closed loop recurrent neural network controller was optimized simultaneously with the morphological parameters using a fixed length genetic algorithm. The results suggested that controller and morphology should co-evolve to produce fitter robots, as is the case in nature [10].

Sims created a system that gives evolution more freedom, where virtual robots compete in a physically simulated 3D world to gain control over common resources. The robots were made of 3D cubes and oscillators [11]. Then, Lipson and Pollack explored the same concept using lower-level building blocks and no sensors. The control was composed of neurons and the morphology was composed of bars and linear actuators. The resulting solutions were remarkably elaborate and difficult to design using traditional methods [12]. Thereafter, Lund investigated the co-evolution of robotic control and morphology using LEGO parts to construct the evolved morphology and downloaded the evolved control to LEGO MINDSTORM RCX [13]. The search space for morphology was limited, but the solution search space was enlarged when co-evolving control and morphology [13, 15].

An obvious constraint on evolution is the manufacturability of resulting solutions. Therefore, Faíña et al. proposed the use of modular robots as the fundamental building blocks for evolutionary processes, because modularity allows building a wide variety of robotic structures, simplifies the search space, and ensures easy implementation in reality [4].

## 3  Soft Modular Robotics

Soft modular robots are composed of various units or modules, hence the name. Each module involves actuators, sensors, computational, and communicational capabilities. Usually, these systems are homogeneous where all the modules are identical; however there could be heterogeneous systems that contain different modules to maximize versatility [4].

Soft modular robotic systems have three promises: versatility, robustness, and low cost. Versatility is the capability of the modular robotic system to form a number of different shapes; each with big numbers of degrees of freedom (DOF). In other words, to allow the robot to self-reconfigure in order to accomplish various tasks in different environments. Versatility can be measured by the number of isomorphic configurations the robotic system can form and by the number of DOF in the system. The number of configurations grows exponentially with the number of modules and the number of DOF grows linearly with the number of modules. Robustness comes from redundancy and self-repair that will be discussed in Section 3.3. When the robot is composed of numerous identical modules and one fails, any other module can replace it to keep the system running. Finally, low cost promise is achieved through batch fabrication. As the numbers of repeated modules increases, the economies of scale come into play and the per-module cost goes down [16]. Also, it can be achieved through rapid prototyping equipment techniques; such as 3D printing, that can build any object by laying down successive layers of material.

Evolutionary robotics can be applied to soft modular robotics to allow self-assembly from constituent modules, self-reconfiguration into different functional forms, self-repair to detect errors recover from failures, and self-reproduction where one system can produce another autonomous functional system.

The literature below is ordered chronologically by publication date for easier sequence demonstration.

### 3.1   *Self-Assembly*

One of the main benefits of modularity is the capability of self-assembly, which is the natural construction of complex multi-unit system using simple units governed by a set of rules. Self-assembly process is ubiquitous in nature as it generates much of the living cell functionality [17]. However, it is uncommon in the technical field, because it is considered as a new concept relatively in that arena although it could help in lowering costs and improving versatility and robustness; which are the three promises of soft modular robotics. The ability to form a larger stronger robot using smaller modules allows self-assembled robots to perform tasks in remote and hazardous environments.

In other words, self-assembly is the problem of designing a collection of elements with edge binding properties such that, when they mix randomly, they bind to form desired assemblies. The elements may be homogenous or heterogeneous; their binding properties may be fixed or dynamic; and they may have a range of capabilities such as ability to detect binding events or exchange information with neighbours [18].

Jones and Mataric in 2003 introduced Intelligent Self-Assembly (ISA) system using Assembly Agents (AA) and Transition Rule Set (TRS) compiler, which takes a goal shape as an input and gives a set of rules as an output that can be utilized by the AAs to assemble the target shape. AAs could be modules in a modular robotic system. Each AA has limited and local sensing and local rule-based control. Increased computational capabilities allows for better interaction among AAs and for assembling more complex structures accordingly. The proposed algorithm organizes the interactions of AA s through the use of the TRS Compiler. This ISA algorithm can be incorporated into distributed reconfiguration algorithms for lattice based self-reconfigurable robots [19].

Stochastically driven self-assembly 2D systems were studied by White et al. in 2004 as they developed algorithms and hardware for few systems. One system uses square modules with electromagnets that self-assembled into an L-shape and then self-reconfigured into a line. The other system uses triangular modules with swivelling permanent magnets that self-assembled into a line and then changed their sequence within the line. Both systems lack batteries, and the modules only receive power after they connect to the structure being self-assembled. A configuration map is distributed to each unit to allow locally determining which of its free bonding sites to activate in order to form a specific geometry, but this approach may lead to deadlocks. Therefore, an alternative to the previous approach is to temporally moderate the formation such that cavities do not form through layered construction [20].

Tolley et al. extended the abovementioned 2D system to 3D. Their evolutionary approach takes a target function as input and designs a robotic structure as output to achieve that input function. These structures are evolved using a frequency-based representation. Then, the assembly algorithm takes place to plan the assembly of the fittest evolved robot by sampling a graph of all possible paths to the target structure and following those that leave the most options open. For each sample, the assembly problem is solved in a reverse order by beginning with the final structure and removing one valid module at a time to go backwards in order to guarantee the existence of a minimum of one path to a complete final assembly at every assembly stage. However, the modules in this system are unable to move on their own, as they need to circulate in turbulent fluid to accrete onto the structure. This fluidic system could be scaled down to produce micro-scale modules [21].

In 2006, Kelly and Zhang proposed a planar distributed assembly model, in which homogenous assembly agents; i.e. modules, move randomly and asynchronously on a 2D grid of cells, attaching square blocks together to form a target structure; such that an agent can fit within one cell. Assembly starts with a seed block, and then the structure grows outwards from the seed. Assembly rules are stored in an internal lookup table, with each rule specifying a binding configuration that activates an assembly action. The group of assembly rules forms an assembly rule set that is identical for all agents in order to allow each agent of performing a complete assembly task if needed. Similar to Jones and Mataric ISA system [19] mentioned earlier, except that the configuration for each assembly rule must be fully specified, and some small differences to allow assembling a larger class of robotic structures [22].

*3.2  Self-Reconfiguration*

Recently, soft modular robotics has gotten attention from researchers in the robotics field due to their ability to self-reconfigure [23]. Modular self-reconfigurable robots involve various modules that can combine themselves autonomously into a meta-module or a structure that is capable of performing a specific task under certain circumstances [4]. Self-reconfigurability allows these robots of metamorphosis, which in turn makes them capable of performing different sorts of kinematics. For instance, a robot may reconfigure into a manipulator, a crawler, or a legged one [23]. This sort of adaptability enables self-reconfigurable robots to accomplish tasks in unstructured environments; such as space exploration, deep sea applications, rescue missions, or reconnaissance [24].

Yim et al. in 2002 classified reconfigurable robots into three classes of architecture: lattice, chain, and mobile based on how they reconfigure [25]. Then, they added deterministic and stochastic reconfigurations in 2007 [26].

- Lattice architectures have modules that are arranged in a 2D or 3D pattern or virtual grid that can be used as a guide for modules to determine their positions and form the new shape accordingly. All modules remain attached to the main body. When units move only to neighbouring positions within a lattice, planning and control become less complex compared to when units move to any arbitrary position [25]. Moreover, lattice architectures are capable of offering simpler reconfiguration compared to other classes, because control and motion can be executed in parallel [26]. This class has received the most research attention due to its less demanding programming. Lattice-type systems exploit lattice regularity when aligning connectors during self-reconfiguration. This allows for faster/easier self-reconfiguration. However, assuming that all modules conform to the lattice can be problematic for systems with a big number of modules [27]. One example of a lattice-based self-reconfigurable robot is Molecule.

- Chain/Tree architectures have modules that are connected together in a string or tree topology. The serial underlying architecture implies that each chain is always attached to the rest of the modules at one or more points, and the modules reconfigure by attaching and detaching to and from themselves. The chains may be used as robotic arms, legs, or tentacles [25]. Chain architectures are more versatile compared to other architectures due to their capability of reaching any point in space through articulation, but they are more difficult to control and more computationally difficult to represent and analyse [26]. An example of a chain-based self-reconfigurable robot is PolyBot.

  It is important to mention that lattice architecture and chain architecture do not contradict, and numerous systems can be both at the same time, such as M-TRAN and SuperBot [27]. These systems tend to have Hybrid architectures.

- Mobile architectures have modules detach from the main body and manoeuvre independently using the environment; e.g. liquid or outer space, to link up at new locations in order to form new shapes, complex chains or lattices, or form a number of smaller robots. Mobile architecture is less explored compared to other structures because the reconfiguration difficulty of outweighs the functionality gain [25-26]. One example of a mobile-based self-reconfigurable is CEBOT.

- Deterministic Architectures have modules move directly to their target locations during the self-reconfiguration process. Each unit's location can be known at all times or calculated at run time, such that reconfiguration times are guaranteed. Feedback control is necessary to ensure precise movement. Usually, macro-scale systems are considered deterministic [28].

- Stochastic Architectures have modules move in a 2D or 3D environment using statistical processes; e.g. Brownian motion, which are used to guarantee reconfiguration times. The exact location of each unit is only known when it is connected to the main structure, but the paths taken by those units to move between locations might be unknown. Stochastic architectures are more ideal at micro-scale systems. The environment provides most of the needed energy for moving units around [26].

The following table lists many modular self-reconfigurable robotic systems along with their architectural class.

Table 1. Self-Reconfigurable Robots Class. Table Courtesy of [26]

| System | Class |
| --- | --- |
| CEBOT | Mobile |
| Polypod | Chain |
| Metamorphosing Robot | Lattice |
| Fracta | Lattice |
| Molecules | Lattice |
| PolyBot | Chain |
| I-Cube | Lattice |
| Crystalline | Lattice |
| TeleCube | Lattice |
| CONRO | Chain |
| MTRAN-II | Hybrid |
| Atron | Lattice |

| System | Class |
|---|---|
| Programmable parts | Stochastic |
| YaMoR | Chain |
| Superbot | Hybrid |
| Molecubes | Chain |

## 3.3 *Self-Repair*

The Self-repair is a special type of self-reconfiguration that allows a robot to replace damaged modules with functional ones in order to continue with the task at hand [23]. A self-repair system must have two qualities: the ability to self-modify, and the availability of new parts or resources to fix broken ones. Therefore, soft modular self-repair robots usually consist of redundant modules. Self-repair consists of detecting the failure module, ejecting the deficient module and replacing it with an efficient extra module. Such robots are well suited for working in unknown and remote environments.

Some of the soft modular robotics systems reviewed later in this article – in the Applications section – will be discussed in terms of self-repair capabilities.

## 3.4 *Self-Reproduction*

The ultimate form of self-repair is self-reproduction; which allows robots to reproduce themselves from an infinite supply of parts using simple rules. If the resulting system is an exact replica of the original, the system is called a self-replicator [29]. The effort in self-reproducing is focused on the design and construction of a small seed system that will grow exponentially to form a larger system through tens of generations. The resulting self-reproducible robots are capable of accomplishing very large-scale tasks; such as collection of solar energy, direct removal of greenhouse gases from the Earth's atmosphere, and water purification for irrigation. Self-reproduction differs from automatic manufacturing or self-assembly, because the resulting systems do not need to make copies of themselves in the latter cases. Since any replication process requires an external material supply, some lattice positions may act as dispensers, where new modules reappear when removed from that location. Self-replication is classified to the following types [28].

- Direct reproduction: A robot picks modules from a dispenser and places them in a new location to gradually build a copy of itself from the ground up.
- Multi-parent reproduction: Multiple robots produce a single copy; such that one machine places modules, while the other assembles these modules.
- Self-assisted reproduction: The robot being built self-reconfigures to assist its own building during the building process.
- Multi-stage reproduction: Temporary scaffold is needed in order to build the target robot. Then, this temporary scaffold is either discarded as waste or re-used to produce additional robots.

Von Neumann was the first to prove the possibility of self-reproduction in 1966 in his close to physical implementation kinetic model of self-reproducing automata, where he aimed to explore computing devices analogous to human brain in which the memory and processing units are tremendously parallel and are capable of repairing and building themselves given the required raw material. Neumann followed Ulam suggestions in [30] to visualize a discrete system comprising a 2D lattice of a finite number of state machines, called cells, interconnected locally. This system can evolve in discrete time steps, so each cell can compute its new internal state. The fitness function is identical for all cells and is a function of the states of the neighbour cells [31]. Today, this system is known as a Cellular Automaton (CA). This research on self-replicating CA was continued later by other authors [32-33].

Chirikjian et al. introduced a concept for self-replicating robotic systems composed of mobile robots, materials processing unit, solar panels and a rail gun. Initial hardware prototypes were constructed from LEGO Mindstorm kits along with enhanced electrical connections and magnetic alignments to demonstrate direct replication. LEGO Mindstorm kits were used to reduce the complexity because of their modular nature and ease of use. Two prototypes were built: Fixture-Based Design and Semi-Autonomous Replicating System. The first prototype is a remote-controlled robot that is not autonomous but can produce a replica of itself. In this design, several passive fixtures are located in the assembly area to assist the robot to assemble a replica of itself. The second prototype is unable to make copies of itself directly. Therefore, the robot makes intermediate systems with different properties than itself. Then, those intermediates can assist the original robot in manufacturing replicas of the original. This prototype system is based on the first prototype results in remotely controlled robotic replication with additional features that enable the robot to perform many subtasks in the replication process autonomously. Although this system is not fully autonomous self-replicating, it is considered as a major stepping-stone in that field [34].

More recently, Griffith et al. demonstrated that self-assembling systems can self-replicate if the intelligent modules were configured to duplicate. The system can self-replicate by selecting the appropriate building blocks from parts distributed in the environment. Is also can self-repair the errors occurred during the copying process. This process enables systems to generate exponential numbers of accurate replicas as a function of time [35].

## 4  Applications

There is a growing number of soft modular robotics prototypes that has been studied in the literature, so in this section we survey a number of emphasized prototypes that participated in the growth of modular robotics research.

The timeline we covered in this paper ranges from 1990 until this year 2016. Figure. 1 illustrates a chronogram of some of the surveyed systems. Tables 2-6 compare some of the surveyed systems based on a number of different parameters.

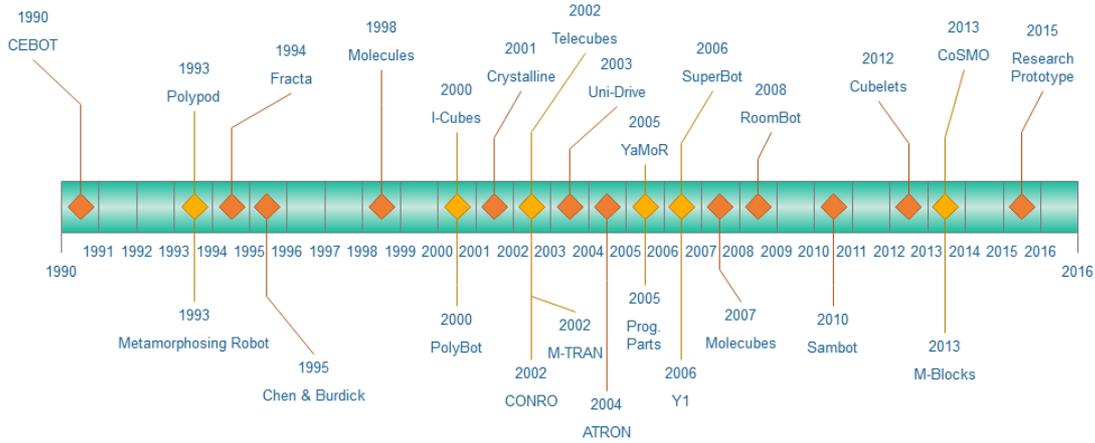

Figure. 1. Chronogram of selected soft modular robotic prototypes

Table 2. Geometrical characteristics of various soft modular robotic systems. Table Courtesy of [27]

| System | Dimensions | Actual DOF | Connectors (Actuated) | Lattice | Geometry |
|---|---|---|---|---|---|
| Metamorphosing Robot | 2D | 3 | 6 (3) | Hexagonal | |
| Fracta | 2D | 0 | 6 (3) | Hexagonal | |
| Molecules | 3D | 4 | 10 (10) | Cubic | |
| PolyBot | 3D | 1 | 2 (2) | Cubic | |
| I-Cubes | 3D | 2 | 2 (2) | Cubic | |
| Crystalline | 2D | 1 | 4 (2) | Square | |
| Telecubes | 3D | 1 | 6 (6) | Cubic | |
| CONRO | 3D | 2 | 4 (1) | None | |

| System | Dimensions | Actual DOF | Connectors (Actuated) | Lattice | Geometry |
|---|---|---|---|---|---|
| M-TRAN | 3D | 2 | 6 (3) | Cubic | 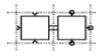 |
| ATRON | 3D | 1 | 8 (4) | Surface-Centered Cubic | 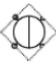 |
| SuperBot | 3D | 3 | 6 (6) | Cubic | 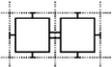 |

Table 3. Electrical characteristics of some soft modular robotic systems. Table Courtesy of [27]

| System | CPU | Power | Communication | Sensors |
|---|---|---|---|---|
| Polypod | Motorola MC68HC11 | Yes | Optical & electrical | Joint position, docking aid, force |
| Fracta | Z80 | No | Optical | None |
| Molecules | None | No | None | None |
| PolyBot | Motorola PowerPC 555 | Yes | Optical & electrical | Joint position, docking aid, orientation, force |
| I-Cubes | PIC 16C63A/73B[c] | Yes | Electrical | Joint position |
| Crystalline | Atmel AT89C2051 | Yes | Optical | Joint position |
| Telecubes | - | No | Optical | Docking aid |
| CONRO | Basic Stamp 2 | Yes | Optical | Docking aid |
| M-TRAN | 3×PIC, 1×TNPM | Yes | Electrical | Joint position, orientation |
| ATRON | Atmel MEGA128L | Yes | Optical | Joint position, orientation and proximity |

Table 4. Physical Characteristics of some soft modular robotic systems. Table Courtesy of [27]

| System | Weight (g) | Dimensions (cm) | Connector Type | Unisex |
|---|---|---|---|---|
| Metamorphosing Robot | - | - | Mech. Hooks | No |
| Fracta | 1200 | ø12.5 | Electro Magnets | No |
| Molecules | - | - | Mech. Hooks | No |
| PolyBot | 200 | 5x5x5 | Mech. Pin/Hole, SMA | Yes |
| I-Cubes | 200 | 6x6x6 | Mech. Lock | No |
| Crystalline | 375 | 5x5x18 (contracted) | Mech. Lock | No |
| Telecubes | - | 6x6x6 (contracted) | Switching Perm. Magn | Yes |

| System | Weight (g) | Dimensions (cm) | Connector Type | Unisex |
|---|---|---|---|---|
| CONRO | 115 | 10.8 × 5.4 × 4.5 | Mech. Pin/Hole, SMA | No |
| M-TRAN | 400 | 6 × 6 × 12 | (versions I&II) SMA+Perm Magnets, (version III) Mech. Hooks | No |
| ATRON | 850 | ø11 | Mech. Hooks | No |

Table 5. Soft modular robotic systems classification based on holistic system characteristics

| | Self-Assembly | Self-Reconfiguration | Self-Repair | Self-Replicate |
|---|---|---|---|---|
| CEBOT | √ | √ | √ | |
| Polypod | | √ | | |
| Metamorphosing Robot | | √ | | |
| Fracta | √ | | √ | |
| Chen & Burdick Robot | √ | √ | | |
| Molecules | | √ | | |
| PolyBot | | √ | | |
| I-Cubes | | √ | | |
| Crystalline | | √ | √ | |
| Telecubes | | √ | | |
| CONRO | | √ | | |
| M-TRAN | | √ | | |
| Uni-Drive | | | | |
| ATRON | | √ | | |
| Programmable Parts | √ | | | |
| YaMoR | | √ | | |
| Y1 | | | | |
| SuperBot | | √ | | |
| Molecubes | | Manually reconfigurable but the replicas can self-reconfigure | | √ |
| RoomBot | √ | √ | | |

|  | Self-Assembly | Self-Reconfiguration | Self-Repair | Self-Replicate |
|---|---|---|---|---|
| Sambot | √ | √ | | |
| Cubelets | | | | |
| M-Blocks | √ | √ | | |
| CoSMO | | √ | | |
| Research Prototype | | | | |

Table 6. Soft modular robotic systems classification based on modularity state of matter

|  | Homogeneous | Heterogeneous |
|---|---|---|
| CEBOT | √ | |
| Polypod | | √ |
| Metamorphosing Robot | √ | |
| Fracta | √ | |
| Chen & Burdick Robot | | √ |
| Molecules | √ | |
| PolyBot | √ | |
| I-Cubes | | √ |
| Crystalline | √ | |
| Telecubes | √ | |
| CONRO | √ | |
| M-TRAN | √ | |
| Uni-Drive | | |
| ATRON | √ | |
| Programmable Parts | √ | |
| YaMor | √ | |
| Y1 | √ | |
| SuperBot | √ | |
| Molecubes | √ | |
| RoomBot | √ | |
| Sambot | √ | |
| Cubelets | √ | |
| M-Blocks | √ | |
| Senior Project | | √ |

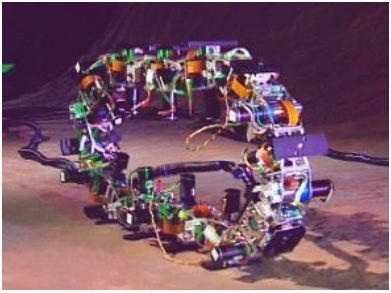
(a)

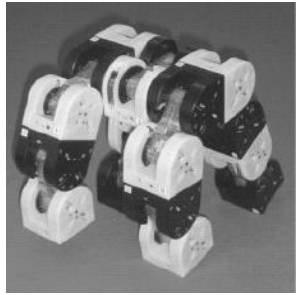
(b)

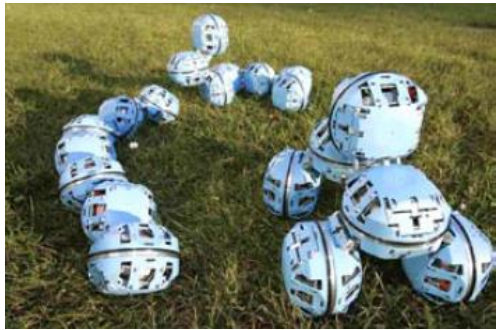
(c)

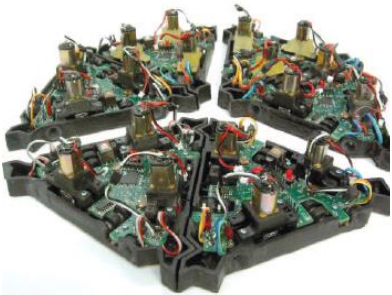
(d)

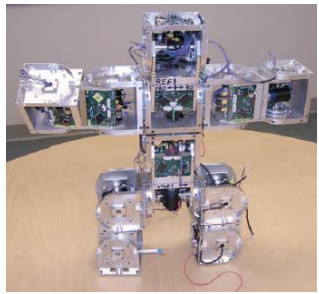
(e)

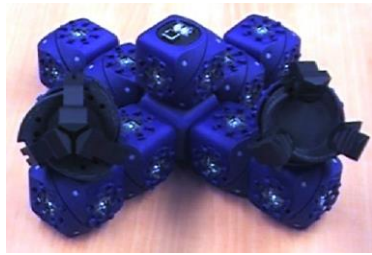
(f)

Figure. 2. (a) PolyBot G2 [53] (b) M-TRAN III [62] (c) ATRON [27] (d) Programmable Parts [26] (e) SuperBot [26] (f) Molecubes [69]

Figure. 2 demonstrates some of the built systems. The subsections below are chronologically ordered by publication date.

### *4.1  CEBOT – 1990*

CEBOT is one of the first modular robots that were developed by Fukuda and Kawauchi in 1990, as a distributed intelligent system. CEBOT is a cellular Dynamically Reconfigurable Robotic System (DRRS) that consists of units called "cells". Those cells can build up modules that connect to other modules to form very complex systems. In addition, these cells can automatically communicate, attach, and detach to perform a function, which allows the system to self-assemble and self-repair [24, 36-37].

CEBOT self-assembly method is designed for a small homogeneous local system that consists of around 10 units. Those units are connected in an arbitrary shape and one unit is chosen to be the origin of construction or the kernel. The kernel gathers adjacent units to compose a logical connection network according to the embedded plan. This network is the first stage. The units involved in the first stage network then gather some surrounding units and form the second stage network. Repeating this process increases the stages, and the network grows stage by stage, approaching the target configuration. The difficulty in construction is low due to using the layer, which acts as a kind of coordinate system to reduce the volume of search spaces [38].

Self-repair can be performed in CEBOT with a simple procedure due to the layered structure of the system. The strategy is to transport spare units to the area of the damage and refill it. This self-repair can be performed by degeneration of the system to the previous stage. The proposed self-repair method consists of 3 steps; Failure Detection, Degeneration Signal, and Degeneration. If several failures occur in the system, signals of several levels are spread, and the system goes back to the lowest level. It must be noted that if the kernel is removed, the units must begin again with the kernel selection process. The authors considered only the simplest case that several units are removed from the system and remained units work correctly. Another simple failure is halting failure; in which failed units do nothing. The proposed method can be applied to this case after cutting off the failure units from the system [38].

## *4.2 Polypod – 1993*

Polypod is a unit-modular system composed of two different modules, the segment and the node. The segment provides 2 DOF, and the node supplies power to the segments. Those segments can self-reconfigure to form different shapes and produce different locomotion gaits accordingly; such as cartwheel, slinky, rolling-track, earthworm, and caterpillar. Additionally, Polypod introduced new robot land locomotion modes; such as two and three-dimensional locomotion gaits and exotic gaits, using a control scheme that combines a small number of primitive control modes for each module [39-41]. PolyBot was the first self-reconfigurable system that demonstrated transitioning from a loop gait into a snake-like gait in 1998. The self-reconfiguration task was accomplished by disconnecting one connection port without any docking [42].

## *4.3 Metamorphosing Robot – 1993*

Soon after Polypod, Chirikjian proposed a dynamically reconfigurable unit-modular robot called Metamorphosing Robot. The mechatronic modules in this system can connect, disconnect by rolling over adjacent modules to allow autonomous self-reconfiguration. Each module is a planar hexagonal shaped robot with 3 DOF with each side of the hexagon capable of connecting to another hexagon of the opposite polarity. Each module allows power and information to flow through itself to its neighbours. As the number of modules in metamorphic system approaches infinity, the manipulator can be viewed as a "mechatronic amoeba" because the manipulator takes on a continuous appearance [41, 43-45]. Later in 2001, Chiang and Chirikjian introduced a cost function to measure reconfiguration fitness and to bisect shapes. This can be viewed as a geometric figures pattern-matching problem under rigid body motions [46].

## *4.4 Fracta – 1994*

Murata et al. designed a 2D robotic system called Fracta in 1994 as a modular robot that is composed of homogeneous mechanical units. Each unit is called a Fractum and considered as the atom of machine. The Fracta system is capable of self-assembly since each unit can connect to other units autonomously to form a given target shape through a diffusion-like process. Each fractum has the potential to become any part of the system and has information about the final shape of the whole system, so it can communicate with neighbouring fracta in order to recognize the local connection and organize the whole shape accordingly. The function of self-assembly has been verified by computer simulation [47].

Then, this work was extended by Yoshida et al. in 1999 to a 3D self-repair system. Self-repair, in this case, was considered as an extension of self-assembly that can detect damage and let the whole system reconstructs itself accordingly. Self-assembly and self-repair were implemented using identical software on each unit with local inter-unit communication. A major difficulty of developing 3D self-assembly algorithm lies in the multiplicity of DOF compared to 2D systems that have to choose only one of two directions, clockwise or counter-clockwise. This algorithm was implemented in a distributed manner to avoid premature convergence to undesired shapes using a stochastic relaxation process based on simulated annealing. A hardware system composed of 20 mechanical units was used for validation [48].

## 4.5  Chen and Burdick – 1995

In 1995, Chen and Burdick introduced a modular robotic system consisting of joint and link units. The joint modules are revolute, prismatic, helical, or cylindrical. The link modules come in two shapes; square prisms with 10 ports or cubic box units with 6 ports. Joint modules are connected to the link modules through connecting ports. The link modules have symmetrical geometry and symmetrically located connecting ports in order to allow link modules to be re-oriented without altering the robot kinematics. The developed robot is capable of self-assembly and self-reconfiguration into a number of different kinematic configurations to solve a given problem [49].

The problem of finding an optimal module assembly configuration for a specific task was solved by a discrete optimization procedure based on assembly incidence matrix representation of the modular robot. Genetic algorithms (GA) were employed to solve this optimization problem, and a canonical method was introduced to represent a modular assembly in terms of genetic strings. However, in some instances, this procedure can be computationally expensive. Therefore, a discrete combinatorial optimization algorithm can be an alternative. In short, GA method is well suited for modular robotic assembly problems. This system can be used with heterogeneous modular robots as well [49].

## 4.6  PolyBot – 2000

PolyBot is a modular self-reconfigurable robot that was implemented by Yim et al. in 2000 to explore how realistic is to implement robots using several homogeneous hardware modules. Three generations of PolyBot modules were prototyped; such that each generation addresses a number of shortcomings discovered in the previous generation. The first generation (G1) is constructed from two module types: nodes and segments. The segments are nominally rectangular prisms and have 1 rotational DOF separating two connection ports. The node modules are fixed passive cubes with six connection ports. Unlike its G1 predecessor, the second generation (G2) connection ports have electromechanical latches under software control. These latch onto the pins protruding from the opposite face. An IR ranging system permits closed loop docking as will be elaborated on in this section. The third generation (G3) modules are smaller and lack the DC motor extending past the side of each module. The new module has instead a DC pancake motor with a harmonic gear that is completely internal. The connectors are larger pitch and have higher contact force for higher current loads to enhance performance.

The first two generations of PolyBot prove versatility by executing locomotion over a variety of terrain. However, as the number of modules increases, cost increases, and robustness decrease due to software scalability and hardware dependency issues. Currently the maximum number of modules utilized in one connected PolyBot system is 32 with each module having 1 DOF. The third generation deals with 200 modules to show a variety of capabilities, including moving like a snake, lizard or centipede as well as humanoid walking and rolling in a loop [50-53].

PolyBot is capable of self-reconfiguration by changing its geometry and locomotion mode depending on the terrain type; rolling over flat terrain, earthworm to move around obstacles, and a spider to step over hilly terrain. Planning the self-collision-free motions can be challenging as the size of this space is exponential in the number of modules, *n*, but proportional to the number of DOF. For many applications, a fixed set of configurations is sufficient. In this case, reconfigurations can be pre-planned off-line and stored in a table for ease of reconfiguration [16].

The same team introduced in [51] PolyKinetic, a system for programming modular self-reconfigurable robots that supports a range of paradigms from posable programming to behavioural coordination. The PolyKinetic software environment consists of an XML-based robot scripting language called PARSL (Phase Automata Robot Scripting Language), a PolyBot simulator, and a programming environment. PARSL allows users to define robot configurations through module groups and their associated sensors and actuators. It also permits definition of gait control tables and automata, and applies these to the module groups. This shifts the focus away from low-level implementation to high-level gait specification. The PolyBot/Polykinetict System is an effective platform for robotics education. PolyBot modules are simple, robust and easy to assemble. The PolyKinetict programming System allows users of diverse skill levels to develop control programs for modular robots.

## 4.7   I-Cubes – 2000

Ünsal and Khosla introduced in 2000 I-Cubes, a 3D modular self-reconfigurable robotic system. I-Cubes is a bipartite collection of individual modules that can be independently controlled. The group consists of active elements, called links, which are 3-DOF manipulators capable of attaching to/detaching from the passive elements (cubes) acting as connectors. The cubes can be oriented and positioned by the links. Using actuation and attachment properties of the links and the cubes, the system can self-reconfigure to adapt to its environment. The links are actuated using servomotors and worm gear mechanisms. Mechanical encoders and rotary switches provide position feedback for semi-autonomous control of the system. The cubes are equipped with a mechanism that provides inter-module attachment [70].

## 4.8 Crystalline – 2001

In 2001, Rus and Vona developed Crystalline distributed robotic system that consists of 3 DOF atoms, which allows expansion and contraction by a factor of two. Robots are formed by expanding and contracting each atom frame in order to move relatively to the other atoms. These movements simulate muscles actuation mechanism which permits automated shape metamorphosis. Moreover, Crystalline robots are capable of self-reconfiguration very fast in $O(n^2)$ time, where $n$ is the number of atoms. These robots carry a number of redundant atoms on their bodies to allow self-repair by ejecting the bad atom and replacing it with a fresh one of the extra atoms [54-57].

Crystalline is capable of self-reconfiguring by assuming any arbitrary geometric shape in a dynamic fashion. Crystalline module motion is controlled by attaching one atom to a neighbouring Atom and actuating the expansion or contraction mechanism. An individual atom cannot relocate without help. However, by contracting and expanding a group of modules in a coordinated way, Atoms can move relative to a structure. Unlike other modular robots, where modules can relocate by traveling on the robot surface, Crystalline atoms can relocate by traveling through the volume of Crystal on a concave structure [54-55].

Fitch et al. built on the work of Yoshida et al. in [48] to accomplish self-repair using Crystalline robot with a focus on geometric motion planning. Crystalline robots can self-repair using a three-phase process: failure detection, failed module ejection, and replacing the failed module with a good one. The authors did not address detecting module failure, because it depends on the system implementation. In order to eject a "dead" module, the "live" modules move it to the ejection position. For that reason, the system should identify all locations on the robot surface where it is possible to eject the dead module, and then compute the shortest path to that location and push the dead module along the shortest path. To improve scalability, the authors developed find-cliffs algorithm to analyse the geometric shape of the robot rather than the number of modules. They also developed an algorithm for moving the failed module to the cliff edge and replacing it with a spare. Self-repair was experimented in simulation and the proposed algorithms support 2D models only [56].

## *4.9   Telecubes – 2002*

Telecubes are cubic modules that were introduced by Suh et al. in 2002, as an extension to the Crystalline system mentioned above. Each cube has 6 prismatic DOF and sides capable of expanding more than twice its original length. Those cubes can form a modular self-reconfigurable robot by attaching and detaching magnetically to other cubes [58-59].

When it comes to reconfiguration, it is assumed the initial and final configurations overlap by at least one meta-module. A module is selected based on the minimum Manhattan distance to begin moving. Then, a route is planned for that selected module using a technique similar to the PacMan algorithm. Once the path is generated, it can be converted into a sequence of motion commands that can be executed. During execution, the meta-modules are divided into active and passive groups. The active modules initiate the planning sequence. The passive modules follow the orders given by active modules to move. This reconfiguration algorithm lacked local decision making and parallel execution [59].

## *4.10  M-TRAN – 2002*

M-TRAN (Modular Transformer) is a distributed lattice-based self-reconfigurable modular robotic system that can metamorphose into various configurations; such as a legged machine generating walking motion. In order to drive M-TRAN hardware, a series of software programs has been developed including a kinematics simulator, a user interface for designing configurations and motion sequences, and an automatic motion planner [60].

M-TRAN II is the second prototype where many improvements took place to allow versatile whole body motions and complicated reconfigurations. Those improvements contain reliable attachment/detachment mechanism, high-speed inter-module communication, on-board multi-computers, accurate motor control, and low energy consumption. The software has been improved as well to verify motions in dynamics simulation and to design self-reconfiguration processes [61].

The third prototype, M-TRAN III, has been developed, with an improved connection mechanism. Various control modes including single-master, globally synchronous control and parallel asynchronous control are made possible by using a distributed controller. Self-reconfiguration experiments using up to 24 units were performed by centralized and decentralized control. Finally, system scalability and homogeneity were maintained in all experiments [62].

M-TRAN changes its configuration by changing the modules positions and connections. However, changing the posture of one module is difficult in some cases, as it involves two modules in cooperation and this makes the problem more complicated. To cope with such difficulty of planning, two types of software have been developed. The first is a motion design interface, which helps a human programmer to design a reconfiguration sequence and motion generation through a powerful graphic interface. The second is a locomotion planner for an M-TRAN cluster, in which the above difficulties are relaxed by introducing some regularity into the structure. The planner for locomotion with reconfiguration enables a serial collection of four module blocks to move along a desired 3-D trajectory through self-reconfiguration. An important issue that has to be addressed in the M-TRAN project is how to design the target configuration itself using an algorithm to generate an optimal or near-optimal configuration for the given task or environment [60].

## *4.11   Uni-Drive - 2003*

In 2003, Karbasi et al. presented a new design for modular serial robot that is composed of lighter modules based on the uni-drive concept. The main element of a uni-drive modular robot is a mechanical drive capable of providing a variable bi-directional speed from a constant uni-directional input velocity. The configuration of the new robot can be changed by the order and number of links and joints that make up each module. In order to reduce the module's weight in the new design, the actuator was replaced with a pair of clutches, because actuator often contributes a significant portion of the module weight [71].

## *4.12   ATRON - 2004*

Another modular self-reconfigurable robot is ATRON, a lattice-based system consisting of approximately spherical modules, where each sphere is constructed as two hemi-spheres joined by an infinite revolute joint. Actuation is realized as rotation around an axis diagonally through the sphere. This design allows for a very stable construction around the actuated joint since a relatively large area is available for mechanics. However, the spherical basic module design makes it hard to have big flat surfaces connecting to each other. With spherical modules, connectors need to establish essential point-to-point contacts between modules, which is not desirable due to high collision probability. Despite that ATRON modules are minimalistic because they have only one actuated DOF, the group of modules is capable of self-reconfiguring in three dimensions [27-72].

### *4.13 Programmable Parts– 2005*

In 2005, Bishop et al. built triangular programmable parts, which can be assorted on an air table by overhead oscillating fans to self-assemble various shapes according to the mathematics of graph grammars. The modules can communicate and selectively bond using mechanically driven magnets, without global knowledge of the full shape. Despite planning to build approximately 100 parts, only six parts were built for design simplicity reasons. Those six parts were used in an experiment that showed these parts react similarly to chemical systems [63]. Then, Napp et al. added kinetic rate data measurements to the previous work of graph grammar in order to yield a Markov Process model [64].

### *4.14 YaMoR – 2005*

YaMoR (Yet another Modular Robot) was presented by Moeckel et al. in 2005. The basic YaMoR module consists of an FPGA and a micro-controller for high computational power needs. Those modules communicate wirelessly using Bluetooth interface, which allows controlling a robot from a computer as well. YaMoR can have multiple configurations such as wheel or caterpillar to support different types of locomotion [73-75].

### *4.15 Y1 – 2006*

Gonzalez-Gomez et al. developed three minimal configurations using only two and three Y1 modules. Each of these modules has 1 DOF and they are capable of attach and detach. Y1 module design is inspired by Polybot G1 modules. Then, they used eight Y1 modules to build a modular worm-like robot, named Cube that is capable of moving in a straight line using a wave propagation gait [76].

### *4.16 SuperBot – 2006*

SuperBot is a multifunctional network of modules that can perform as both lattice-based and chain-type self-reconfigurable robots. It was developed by Salemi et al. in 2006 to enhance the mechanical design of M-TRAN, mentioned earlier, by adding an additional rotational DOF between the two existing rotation axes resulting in 3 DOF per module. Each module consists of three main parts: Two end effectors and a rotating central part. SuperBot was designed to be a flexible, strong, and durable robot that can be used in real world applications and also to support multi-mode locomotion [77].

### *4.17 Molecubes – 2007*

Molecubes is an open hardware and software platform for modular robotics that was developed in 2007 to remove entry barriers to the field and to accelerate progress. The system is composed of homogeneous modules with one rotational DOF. Different types of active modules; such as gripper, actuated joint, controller, camera, and wheel along with a number of passive modules were presented. Each robot is a cube shaped with round corners and comprises approximately two triangular pyramidal halves connected with their bases so that their main axes are coincident. These cube halves are rotated by the robot motor about a common axis relative to each other. In this way, every module has one actuated DOF. Each of the six faces of a robot is equipped with an electromechanical connector that can be used to join two modules together. Symmetric connector design allows 4 possible relative orientations of two connected module interfaces, each resulting in different robot kinematics. Evolutionary search was used to design different types of robots rapidly [69, 78].

### *4.18 RoomBot – 2008*

RoomBot is a modular robot that can self-assemble and self-reconfigure into different pieces of furniture. It introduces passive elements in the robot structure, the implementation of a Central Pattern Generator for generating the command of the motors, and the possibility of use a motor in oscillation and constant rotation [79].

### *4.19 Sambot – 2010*

Sambot is a mobile self-assembly modular robot that was implemented by Wei et al. in 2010. Several modules can self-assemble to form a particular structure through a 4-phase autonomous docking process. Also, the resulting shape can reconfigure into different structures that are capable of locomotion. Each module includes an active docking interface and an autonomous mobile body. A pair of detecting infrared sensors is installed on the autonomous mobile body to detect obstacles in front of the robot. In addition, a pair of approaching infrared sensors is also installed on the autonomous mobile body to monitor the relative positions of the modules and provide navigating information for the docking. The computing platform provided for each module is distributed and consists of a number of interlinked microcontrollers. The interaction and connectivity between different modules is achieved through infrared sensors and Zigbee wireless communication in discrete state and control area network bus communication in robotic configuration state. [80].

### *4.20 Cubelets – 2012*

Modular robotics, a robotics construction kit known commercially as Cubelets was presented by Schweikardt and Gross as an educational tool composed of several modules that snap together to construct robots. The modules include various cubes with specific actuation (drive, rotation), communication (light, sound), sensing (distance, temperature, knob, brightness), and computation (min, max, inverse) capabilities, as well as structural parts (blocker, passive, battery). Cubelets exchange sensor information and allow the construction of simple autonomous robots [81-82].

### *4.21  M-Blocks – 2013*

Romanishin et al. introduced M-Blocks in 2013 as a cubic modular robotic system that is capable of self-assembly and self-reconfiguration. However, this robotic system cannot self-repair due to the lack of intelligence incorporation into the system. The cube shaped modules use pivoting motions to reconfigure and change geometry. The robot has three critical systems: the magnetic bonding and pivoting mechanism, the inertial actuator, and the electronic control system. The magnetic mechanism allows the modules to quickly form magnetic hinges on any of the cubes' twelve edges. The inertial actuator is one-dimensional uni-directional actuator that is not strong enough to execute all lattice moves reliably. The electronics include a custom designed PCB [83].

### *4.22  CoSMO - 2013*

The Collective Self-reconfigurable Modular Organism (CoSMO) is the first triple hybrid (lattice, chain and mobile type) mobile Modular Self-Reconfigurable (MSR) robot that has increased computational capabilities and communication bandwidth between connected modules compared to other MSR robots. The modules can share energy with each other and they can move in the main directions. The architecture involves numerous processes running on the µClinux operating system. The inter-process communication is achieved using SOAP calls that are generated by gSOAP Toolkit. The SOAP messaging is encapsulated by the Heavily Decoupled Multi Modular Robots (HDMR) API Interface developed by the authors. CoSMO was evaluated by a number of tests to demonstrate robustness and flexibility where it produced more than 60 robots [65].

*4.23 Research Prototype – 2015*

In 2015, Liu et al. designed a low-cost reconfigurable modular robotic platform that can be used as a teaching tool or a prototype for research in modular robotics. The base module contains two motors and a micro-controller to run and regulate these motors. Each module casing has magnetic connections to cooperate with other modules and create a moving system [84].

## 5 Current State of the Art

More recently, new efforts have been pursued in the fields of evolutionary robotics, soft modular robotics, and in each of the previously mentioned sub-fields; self-assembly, self-reconfiguration, self-repair, and self-reproduction. Many tasks have been shown to be achievable, especially with the high number of physically implemented robotic systems. The following table classifies the aforementioned modular robotic systems according to the implementation method; in simulation vs physical implementation.

Table 7. Soft modular robotic systems classification based on implementation

|                      | Simulation | Physical Implementation |
|----------------------|------------|-------------------------|
| CEBOT                | √          |                         |
| Polypod              |            | √                       |
| Metamorphosing Robot | √          |                         |
| Fracta               |            | √                       |
| Chen & Burdick Robot | √          |                         |
| Molecules            |            | √                       |
| PolyBot              |            | √                       |
| I-Cubes              |            | √                       |
| Crystalline          |            | √                       |
| Telecubes            |            | √                       |
| CONRO                |            | √                       |
| M-TRAN               |            | √                       |
| Uni-Drive            |            | √                       |
| ATRON                |            | √                       |

|  | Simulation | Physical Implementation |
|---|---|---|
| Programmable Parts |  | √ |
| YaMor |  | √ |
| Y1 |  | √ |
| SuperBot |  | √ |
| Molecubes |  | √ |
| RoomBot |  | √ |
| Sambot |  | √ |
| Cubelets |  | √ |
| M-Blocks |  | √ |
| Senior Project | √ |  |

The majority of the prototyped systems were 3D printed; therefore we discuss 3D printed robots in the next subsection, followed by automatic design and manufacturing.

## 5.1    3D Printed Robots

Robot manufacturing is currently highly specialized, time consuming, and expensive, which results in limiting accessibility and customization. Nevertheless, rapid prototyping techniques; such as 3-D printing, are becoming increasingly accessible due to their low cost and high ability of achieving complex geometries. Therefore, different robotic fields start utilizing these planar fabrication methods in order to create 3D printed robotic prototypes.

Onal et al. proposed a new method, called printable robots that can be used to rapidly fabricate capable, agile, and functional 3D electromechanical machines. The new approach takes advantage of available planar fabrication methods to create integrated electromechanical laminates that are subsequently folded into functional 3D machines employing origami-inspired techniques. To demonstrate this print-and-fold process, several prototypes were created that address the canonical robotics challenges of manipulation and locomotion; such as the robot shown in Figure. 3. This technology can be utilized to create a robot-printing machine that requires no technical knowledge on the part of the user after automating some fabrication steps that were performed manually in the proposed system; such as laminating and fabricating [66].

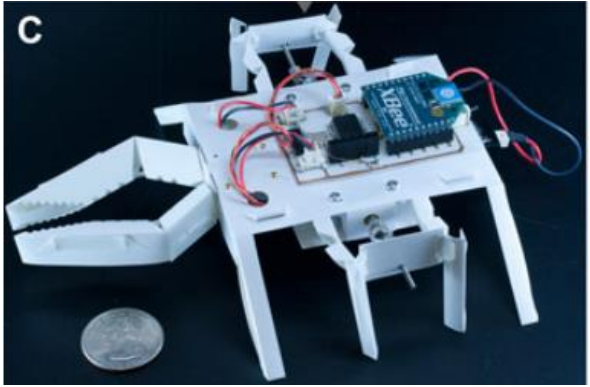

Figure. 3. Origami Inspired Printed Robot [66].

Qi et al. used 3D printing method to fabricate the components of a robotic arm, which provides more precise dimensions and huge time and cost saving in fabrication. The robotic arm is designed with 4 DOF and equipped with 4 servomotors to link the parts and bring arm movement. It is programmed to accomplish accurately simple light material lifting tasks to assist in the production line in any industry [67]. MacCurdy et al. introduced a novel technique for fabricating functional robots using 3D printers. Simultaneously, depositing photopolymers and a non-curing liquid allows complex, pre-filled fluidic channels to be fabricated. This new printing capability enables complex hydraulically actuated robots and robotic components to be automatically built, with no assembly required. The technique is showcased by printing linear bellows actuators, gear pumps, soft grippers and a hexapod robot, using commercially available 3D printer [68].

## 5.2 *Automatic Design and Manufacturing*

Robots automatic design and manufacturing combine evolutionary computation and additive fabrication; such that the former is used for design and the latter for reproduction. The evolutionary computation process operates on candidate robots population to iteratively select fitter machines, create offspring by adding, modifying and removing building blocks using a set of predefined operators, and replace them into the population. Similarly, additive fabrication technology has been developing in terms of materials and mechanical fidelity but has not been placed under the control of an evolutionary process yet.

Lipson and Pollack tried to bridge the reality gap by proposing an approach based on the use of only elementary building blocks and elementary operators in design and fabrication process. Elementary building blocks were used to minimize inductive bias and maximize architectural flexibility. Also, they allow the fabrication process to be more systematic and versatile.

The pre-assembled machine was fabricated as a whole single unit, with plastic supports to connect the moving parts. These supports broke at first motion. Then, standard stepper motors were snapped in, and the evolved neural network was executed on a microcontroller to activate the motors. Three physical machines; shown in Figure. 4, successfully reproduced their virtual ancestors' behaviour in reality [12].

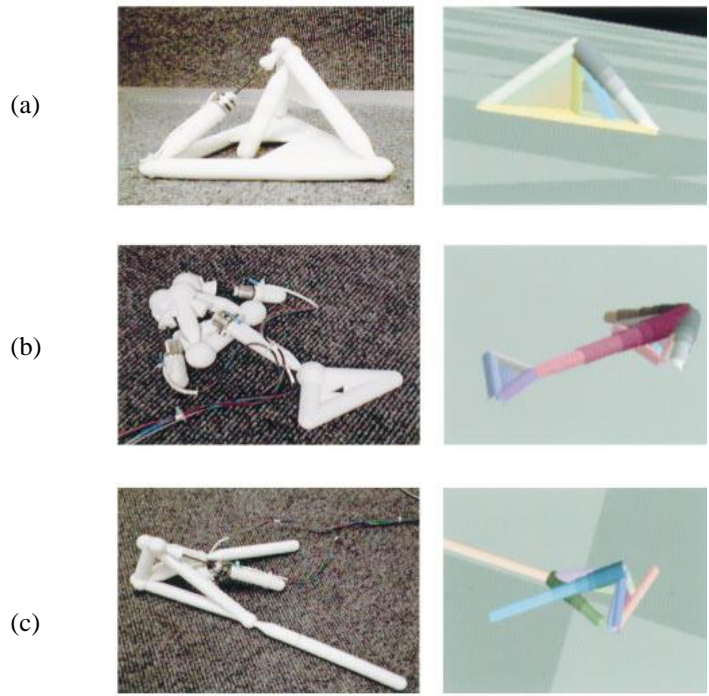

Figure. 4. The resulting robots. Real robots (left); simulated robots (right). (a) Tetrahedron (b) Arrow (c) Pusher [12].

## 6   Conclusion

In evolutionary robotics, reality gap is a big impediment to advancement. Many studies were conducted to cross the reality gap. Conversely, this article surveys the literature in the fields of evolutionary robotics and soft modular robotics to showcase what was accomplished in both fields and how evolutionary robotics can be applied to soft modular robotics to allow self-assembly, self-reconfiguration, self-repair, and self-reproduction. A number of prototypes were discussed in terms of evolutionary techniques and soft modular characteristics. Then, the current state of the art was covered to introduce the new technologies used in the arena including 3D printing and automatic manufacturing.